\def\year{2020}\relax
\documentclass[letterpaper]{article} 
\usepackage{aaai}  
\usepackage{times}  
\usepackage{helvet} 
\usepackage{courier}  
\usepackage[hyphens]{url}  
\usepackage{graphicx} 
\usepackage{graphicx}  
\title{Imagination-Augmented Deep Learning for Goal Recognition}

\author{Thibault Duhamel, Mariane Maynard and Froduald Kabanza\\
 PLANIART \\
 Université de Sherbrooke \\
 \texttt{\{thibault.duhamel, mariane.maynard, froduald.kabanza\}@usherbrooke.ca} \\}

\usepackage{amsthm}
\usepackage{amsmath}
\usepackage{amsfonts}
\DeclareMathOperator*{\argmax}{argmax}
\DeclareMathOperator*{\softmax}{softmax}
\newtheorem{definition}{Definition}
\newtheorem{observation}{Observation}

\newcommand{\citet}[1]{\citeauthor{#1} (\citeyear{#1})}

\begin{document}

\maketitle

\begin{abstract}  
Being able to infer the goal of people we observe, interact with, or read stories about is one of the hallmarks of human intelligence. A prominent idea in current goal-recognition research is to infer the likelihood of an agent's goal from the estimations of the costs of plans to the different goals the agent might have. Different approaches implement this idea by relying only on handcrafted symbolic representations. Their application to real-world settings is, however, quite limited, mainly because extracting rules for the factors that influence goal-oriented behaviors remains a complicated task. In this paper, we introduce a novel idea of using a symbolic planner to compute plan-cost insights, which augment a deep neural network with an imagination capability, leading to improved goal recognition accuracy in real and synthetic domains compared to a symbolic recognizer or a deep-learning goal recognizer alone.
\end{abstract}

\section{Introduction}

Goal recognition is a fundamental cognitive ability, naturally performed by humans during their interactions. Often operating unconsciously, it is a crucial mechanism granting the possibility to foresee and integrate what may happen in the future to make better decisions according to additional projected information, either in cooperative or competitive environments. Artificially intelligent agents, however, still lack such powerful features despite recent breakthroughs in the field.


One of the trending paradigms for implementing goal recognition algorithms relies on plan costs computed by a symbolic planner that inverses the planning process of the observed agent, leveraging the fact that they tend to act rationally towards their pursued goal \cite{ramirez:geffner:10,masters:sardina:19a}. By computing plan cost differences, these methods indicate whether the agent is deviating from an optimal course to the goals and rank them according to this estimation. While promising, these approaches still convey significant challenges before they can be applied successfully to various real-world settings. First, non-trivial alterations are necessary to make reasonable inferences for situations where the optimality of the agents cannot be guaranteed~\cite{masters:sardina:19b}. Second, handcrafted representations used by the symbolic planner to compute expected plan costs may not be complete or precise, and symbolic planners are sensitive to such inaccuracies.

The gist of these approaches is that plan costs are good predictors of the goals pursued by the observed agents. They convey insight about which goals might be more demanding to achieve than others in the future, and require a planning process to derive them. This suggests we could use a deep learning method to learn to predict goals using plan costs as features. From this perspective, a deep neural network equipped with a planner to generate plan-cost features appears to be an imagination-augmented deep neural network~\cite{racaniere:etal:17}. Indeed, an imagination-augmented deep neural network can learn a policy from features generated by an \textit{imagination} module, providing insight about the different futures that may occur if the agent takes any action in a set of possible ones.

Based on this analogy, we developed a novel approach to learn a goal-prediction model from features based on plan costs, with the idea that plan costs will convey insight about the future, improving the accuracy of a deep neural network compared to a baseline not using plan costs. On the other hand, given that plan costs are used as features of a learning algorithm, our hypothesis is that, unlike symbolic cost-based plan recognizers, our learned model would be more robust to errors in the representation used to compute plan costs, without requiring any non-trivial alteration of the inference algorithm to deal with situations where agents are not behaving optimally. We expect the model to automatically learn from data the extent to which an observed agent acts optimally in certain circumstances. We demonstrate the power of this novel idea by implementing two different methods to compute symbolic plan-cost-based features, respectively, \textit{gradients of costs} and \textit{sequential deviations}. We show that each of them enables a deep neural network architecture to learn to better predict the goal of an observed agent than without such plan-cost-based features.

The rest of the paper is organized as follows: first, we provide key background concepts about the problem we solve. Then, we present our method, followed by the setup and results of our experiments. Finally, we provide a brief review of the literature related to our research work.

\section{Background}

Let us give a general definition of a goal recognition problem:
\begin{definition}
A goal recognition problem is a tuple $\langle G, O \rangle$ where $G$ is the set of possible goal states and $O = o_0, \ldots, o_t$ is the sequence of observations of an agent's behavior. $O$ is generated from the interaction of the agent with an environment $E = \langle S,A,c \rangle$, composed of a set of states $S$, a set of actions $A: S \times S$ and a cost function $c: A \to \mathbb{R}^+_0$.
\end{definition}

In this paper, we assume full observability of the agent, i.e. we suppose we can fully extract $E$ from $O$ as well as $s_0, \ldots, s_t \in S$, the sequence of completely observed states, $s_0$ being the initial state. We also use the term \textit{plan} to refer to a sequence of actions $a_0, \ldots, a_t \in A$ pursued by an agent, and $c(s_0, g)$ to define the cost of an optimal plan achieving $g \in G \subseteq S$ starting from $s_0$, where $c(s_0, g) = c(a_0) + c(a_1) + \ldots + c(a_t)$.

In this section, we provide background concepts explaining how to resolve this problem following two paradigms: cost-based goal recognition and goal recognition as learning.

\subsection{Cost-Based Goal Recognition}

The intuition behind cost-based goal recognition is that, assuming that the observed agent is rational (also known as cost-sensitive), they will be more likely to pursue the less costly plans. To perform goal inference, an observer only needs to compare the cost of the observed plan with the cost of an optimal plan for any given goal, computed using an optimal planner over a domain theory of the environment. If the two costs match, then this goal is considered plausible~\cite{ramirez:geffner:09}.

Extending the inference with a probabilistic dimension is a mechanism partially coping for potential divergences from the optimal behavior. For instance, \citet{ramirez:geffner:10} compute the goal inference using a Boltzmann distribution:
\begin{equation}
    P(g|O_{0:t}) = \alpha \frac{1}{1 + \exp(\beta\Delta(s_0,g,O_{0:t}))}
    \label{prob:dist}
\end{equation}
where $\alpha$ is a normalisation factor, $\beta$ is a temperature hyperparameter tuned according to the agent's assessed optimality, and $\Delta$ is the following cost difference formula:
\begin{equation}
    \Delta(s_0,g,O_{0:t}) = c(s_0,g,O_{0:t}) - c(s_0,g,\bar{O}_{0:t})
    \label{costdiff:rg}
\end{equation}
where $c(s_0,g,O_{0:t})$ is the cost of an optimal plan from $s_0$ to $g$ complying with the observed actions in $O_{0:t}$, and $c(s_0,g,\bar{O}_{0:t})$ is the cost of an optimal plan reaching $g$ where at least one of the observed actions has not occurred.

\citet{vered:etal:16} rather use a cost ratio to make a probabilistic inference:
\begin{equation}
    P(g|O_{0:t}) = \alpha \frac{c(s_0,g)}{c(s_0,g,O_{0:t})}
    \label{costdiff:vk}
\end{equation}

\citet{masters:sardina:19a} use a simpler cost difference formula accounting only for the initial and last observations:
\begin{equation}
    \Delta(s_0,s_t,g) = c(s_t,g) - c(s_0,g)
    \label{costdiff:ms}
\end{equation}

This method makes offline costs computing possible for some domains, such as the discrete navigation one, for which we can store the costs into convenient cost maps. They also suppose a Boltzmann probability distribution over this difference.

The ingenuity of cost-based goal recognition lies in the features used to compute goal inferences (optimal plan costs), which are quantities ranking the imagined future according to the agent's rationality.
However, computing a plan, even in the simple case of a deterministic environment under full observability, is NP-Complete~\cite{bylander:94}. These methods cannot be applied realistically in situations where an agent needs to infer the goal of others quickly and where offline storage is not as trivial as in the navigation domain~\cite{masters:sardina:19a}. Approximate plan costs, computed by suboptimal planners or heuristic functions that run faster, can be used to infer an approximate distribution~\cite{ramirez:geffner:09}. They are helpful in situations where the essential matter remains to identify the goals that are more likely.

\citet{vered:kaminka:17} introduced heuristics directly into the goal recognition inference process to judge whether a new observation changes the ranking of goals or whether a goal can be pruned, effectively reducing the number of calls to the planner.

Another work worth mentioning is the one of \citet{sohrabi:etal:16}, which computes the top-$k$ optimal plans for each goal and adds a degree of compliance with the observations to their cost to deal with noisy and missing observations. Those additional quantities make it potentially more robust to suboptimal behaviors and errors in the model, but, in opposition to other works using suboptimal plan costs as predictors, it introduces a significantly higher computation overhead. Indeed, the method computes $k$ times more plans than \citet{ramirez:geffner:10}'s technique, including both the optimal and suboptimal ones, and the value of $k$ must be high to achieve comparable performance.


Other various studies present different ideas to reduce computation times using heuristic metrics instead of plan costs, with reduced accuracy. For instance, \citet{e-martin:etal:15} compute cost interaction estimates in plan graphs, while \citet{pereira:etal:17} use landmarks, with the idea that goals with a higher completion ratio are more likely. We follow this line of inquiry by feeding one of our methods with heuristic metrics as an approximation of plan costs.

\subsection{Goal Recognition as Learning}

Another limit of previously presented methods lies in the inference algorithm, which relies exclusively on symbolic domain knowledge. If the knowledge happens to be incorrect, the algorithms may return inaccurate results. It thus becomes useful to have an adaptive inference process that can account for potential bias in the provided knowledge.

It is where learning algorithms intervene. While plan costs features extracted from domain knowledge still convey bias, the idea is to use an unbiased goal inference algorithm directly from data by automatically extracting patterns from observed examples.

Given a set of goal recognition problems $\langle \mathcal{G}, \mathcal{O} \rangle$, let us assume that there exists an optimal probability function $P$ that is maximal for a true goal $g^*\in G\in \mathcal{G}$, provided with the corresponding observations $O\in \mathcal{O}$, that is, $\argmax_{g \in G} P(g|O) = g^*$.

Given the temporal nature of the sequence $O = o_0, \ldots, o_t$, this probability distribution can be approximated using a recurrent neural network such as a long short-term memory (LSTM) network:
\[ P(g|O) \approx P'(g|O;\theta) = \mathit{LSTM}(O;\theta) = \softmax(h_t)\]
where $\theta$ are the learned parameters of the network, and $h_t$ is a transformation of $O$ recursively defined as $h_t = \tanh(f(o_t, h_{t-1}; \theta))$, where $f$ is a transformation over $o_t$ and $h_{t-1}$ using $\theta$.

Assuming we have access to a training dataset of paired examples $(O, g^*)$ (i.e., we know the true goal $g^*$ for a given $O\in \mathcal{O}$), we can train the set of parameters $\theta$ to minimize the number of erred predictions in our dataset of examples. In other words, we wish to minimize
\[ L=\sum_{n=0}^N l(\mathit{LSTM}(O^n;\theta), g^*_{n}) \]
where $l$ is a loss function (such as the categorical cross-entropy) that is increasingly positive as $P'(g^*|O;\theta)$ approaches $0$.

If the observations are non-symbolic, it can become useful to extract spatial information about the world as well using a spatiotemporal deep neural network (STDNN). In that case, we compute $P'$ in the following manner:
\[ P'(g|O;\theta) = \mathit{STDNN}(O;\theta) = \mathit{LSTM}(O';\theta) \]
where $O' = o'_0, \ldots, o'_t$ is a spatial-wise transformation of $O$ using, for instance, convolutional layers in the case of 2D navigation.

Some works explored LSTM networks trained on observed data for the task of goal recognition~\cite{min:etal:16,amado:etal:18}. However, these networks were trained and applied in single environment domains, and it is not realistic to expect they could generalize to multiple environments.

It is where it becomes handy to explore plan-cost features, providing the model with cross-domain insight about the causal and long-term reasoning necessary to make informed goal inferences.

\section{Method}

We herein present our method as a combination of both paradigms, using neural networks fed by symbolic cost-based predictors. We introduce two novel features and approaches to learn from them.

\subsection{Gradients of Costs (GC)}

Previous works in cost-based goal recognition established that plan costs are reliable predictors. They tend, in fact, to suggest that at least two plan costs are necessary (one derived from the observed plan and another being non-contextual) and seem to withhold useful clues when comparing and assessing the likelihood of the goal.

Considering the cost difference of \citet{masters:sardina:19a} given in equation~\ref{costdiff:ms}, we observe that the cost $c(s_t, g^*)$ decreases as the agent completes its plan, increasing (in the negatives) the difference with $c(s_0, g^*)$. Therefore, it only makes sense that the probability of $g^*$ increases as the difference widens.

In fact, if the behavior of the agent is purely rational, we can make the following observation: 
\begin{observation}
Let $s_0, \ldots, s_n$ be a sequence of observed states and $g^*$ be the true goal of the observed agent. Assuming their plan is optimal, then $c(s_t, g^*) \leq c(s_{t-1}, g^*) \forall t \in [1, n]$.
\end{observation}
Intuitively, the remaining cost of an optimal plan can only monotonically decrease as the agent advances towards their goal.

From this observation, we engineered a novel goal recognition feature by considering the partial derivative of an optimal cost over time. We compute the feature as follows:
\begin{equation}
    \frac{\partial c(s_t, g)}{\partial t} = c(s_{t-1}, g) - c(s_t, g)
    \label{costs_derivative}
\end{equation}
where $c(s_t, g)$ is the optimal cost from the agent's state $s_t$ to $g$\footnote{Since we consider discrete timesteps, we approximate the partial derivative for a single timestep delta. We use the previous point $t-1$ so that the formula does not depend on future information.}. By calculating the derivative for every possible $g \in G$ and every point in time $t \leq |O|$, we obtain a vector of partial derivatives that we define to be the \textit{gradient} of costs (GC):
\begin{equation}
    \mathit{GC}(O) = \bigg[\bigg[\frac{\partial c(s_t, g)}{\partial t}\bigg]_{0 < t \leq |O|}\bigg]_{g \in G}
    \label{grads}
\end{equation}
 
In other words, $\mathit{GC}(O)$ gives a global idea about the current \textit{moving direction} of the agent and which goal states are \textit{towards} their move. Using these as predictors and with an adapted inference algorithm, we can, in fact, obtain a goal recognition algorithm as effective as \citet{masters:sardina:19a}'s to evaluate rational behavior.


The interesting part appears to be its potential to make \textit{better} goal inferences than~\citet{masters:sardina:19a} and \citet{ramirez:geffner:10} for apparent \textit{irrational} behavior. Indeed, making inferences over multiple cost differences instead of a single one saves more information about the observations, hence allowing more flexibility. This process is crucial to keep the system robust against certain misbeliefs conveyed by the domain knowledge.

Indeed, the key in our method is to let the inference algorithm find how to optimally balance past and future information about the agent. If the agent behaves irrationally at first, it may become beneficial to discard the few first observations that do not give useful information about the goals. All the same, this information should not be discarded when they are being rational. Depending on the level of rationality of the agent, our algorithm can find the optimal way to weigh the quantities in time from examples of their behavior, and know when and where to \textit{cut} the past.

\begin{figure}[htb]
    \centering
    \includegraphics[width=\linewidth]{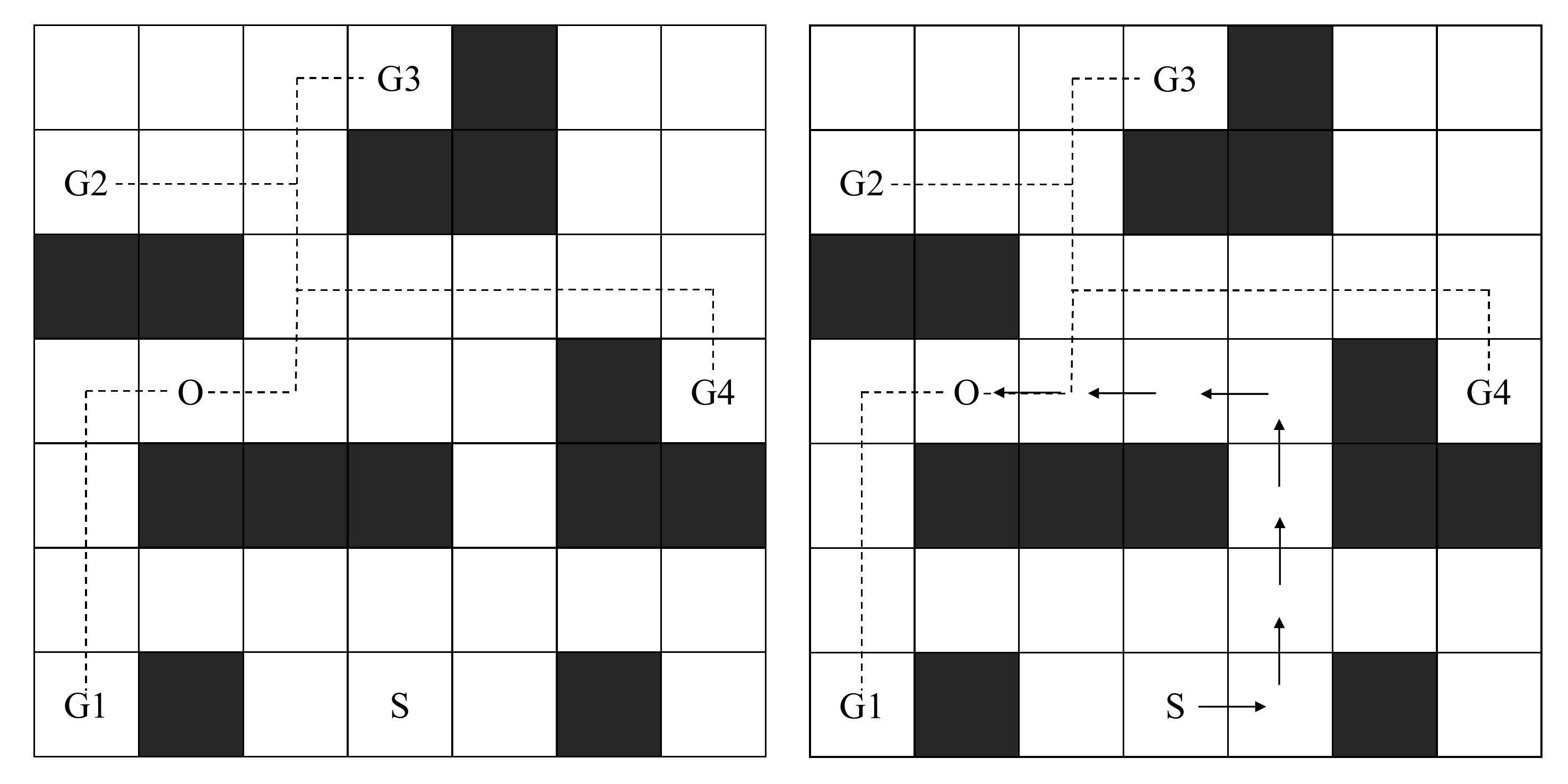}
    \caption{Example of a suboptimal agent navigating in a grid. S is their initial position, O is their last observed position, G1 to G4 are potential goals, and the dashes are imaginary optimal paths. (Left) Without the observations in-between S and O, it is unclear what the destination of the agent is. (Right) With all the observations (arrows), G1 now appears as a likely goal.}
    \label{fig:gridexample}
\end{figure}

Let us consider the motivating example depicted in figure~\ref{fig:gridexample}, where an agent navigates in a specific environment. The agent's behavior is suboptimal for all the goals since $O$ is not on any optimal path to them\footnote{Following the definitions introduced by~\citet{masters:sardina:19b}, the agent is strictly less rational, but not uniformly less rational.}. Yet, this situation could realistically happen, if we imagine that the agent changed their mind, if some paths are less desirable than others, or if there are unseen obstacles. The point is, any misbelief conveyed by the knowledge we have about the world and the agent (deterministic, fully observable, uniform costs) can become problematic when the inference algorithm is fixed precisely over an engineered quantity.

Indeed, the information conveyed by the cost differences of equations~\ref{costdiff:rg},~\ref{costdiff:vk} and~\ref{costdiff:ms} is here ambiguous and lead to counter-intuitive results. For instance, equation~\ref{costdiff:ms} ranks both G2 and G3 first (since equation~\ref{prob:dist} is maximal when $\Delta$ is minimal) followed by both G1 and G4. Since it relies only on two observations to make an inference (as depicted on the left pane of~\ref{fig:gridexample}), crucial information residing in the other observations do not weigh in the decision. Indeed, looking at the right pane and knowing that the agent started a loop, it now seems reasonable to consider G1 as more likely than G4.

Yet interestingly enough, equations~\ref{costdiff:rg} and~\ref{costdiff:vk} make the same prediction, even though their cost formulas rely on all observations. The reason is that they reduce the information conveyed in the observations to only two optimal costs for each goal. On the other hand, GC is computed at multiple points in time and our learned inference algorithm can weigh each gradient feature according to their position in the sequence. Equations~\ref{costdiff:rg},~\ref{costdiff:vk} and~\ref{costdiff:ms} do not allow to do it in the time dimension, since they always compare to the initial projected future at $s_0$, while each quantity of equation~\ref{grads} is function only of two consecutive timesteps. The first values of $GC(O)$ do not affect the latest ones. 
Therefore, though produced by potentially inaccurate domain knowledge, gradients of costs contain all necessary information for our goal inference solution to balance the past and the future.


\subsection{Sequential Deviations (SD): an Approximation of GC}

While costs and gradients of costs convey meaningful information, they rely on expensive planners and a complete model of the environment. We explored the possibility to provide clues to a neural network but, this time, in the form of heuristic functions to lower computation costs.

A heuristic function is a function $h$ that estimates the cost (or distance) of the optimal plan from a start state to a goal state. By extension, it can also take two states as parameters and compute an estimate of the distance between them. In the navigation domain, for instance, the $L_2$ (euclidean) distance is commonly used as a heuristic, since it represents the cost of perfect paths, from a bird's-eye view, for an unconstrained agent. In the rest of this paper, $h$ will denote any heuristic function.

We first approximate the gradients of costs using $h$, such that we obtain a derivative:
\begin{equation}
    \frac{\partial h(s_t, g)}{\partial t} = h(s_{t-1}, g) - h(s_t, g)
\end{equation}

In the general case, the heuristic function does not decrease monotonically along the steps of an optimal path. However, if the heuristic is admissible (never over-estimating the real optimal cost), we can apply the squeeze theorem and conclude that the heuristic will overall converge towards zero.

We introduce the \textit{sequential deviation} (SD) metric, which estimates a temporal deviation of an observed path $O$ to every goal, defined as follows:
\begin{equation}
    \mathit{SD}(O) = \bigg[\Big[\frac{\partial h(s_t, g)}{\partial t}\Big]_{0 < t \leq |O|}\bigg]_{g \in G}
\end{equation}

Although approximating GC, SD still illustrates the global motion of the agent. Looking at the figure \ref{fig:gridexample} again, the $L_2$ heuristic function starts by increasing for G1, but then decreases. An inference algorithm attributing smaller weights to the past values would thus conclude that G1 is likely.

This method shows significant advantages. First of all, it allows bypassing the need for an environment model and planner in specific domains (for which a heuristic independent of the domain is known, such as navigation). Second, it is an approximation of the GC method, hence reducing the computation cost  without losing the generalization capability. Moreover, the sequential deviation metric still encapsulates more temporal information than just the differences of costs from symbolic cost-based approaches.

\section{Experiments}

Although these different paradigms could extend to task-planning problems, the experiments were limited to navigation benchmarks to allow a fair comparison with \citet{masters:sardina:19a}'s state-of-the-art algorithm. We begin by experimenting in a real-world setting, then conduct additional tests on arbitrarily complex navigation settings \cite{masters:sardina:19a}, to compare how incorrect models affect the predictions.

\subsection{Pedestrians on a Crowded Street}

\begin{figure}[htb]
    \centering
    \includegraphics[width=0.8\linewidth]{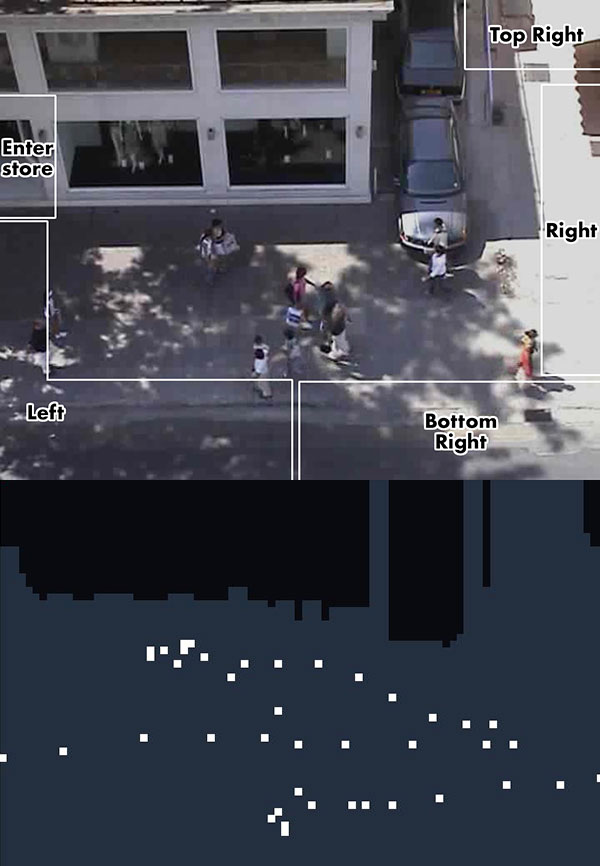}
    \caption{(a) On the top, the different goals achieved by the pedestrians in the video. (b) On the bottom, the grid environment extracted from the raw video. The white dots represent the path of one person (trajectories are not contiguous but sampled evenly). It is clear that the observed behaviors are erratic.}
    \label{fig:ucyimage}
\end{figure}

UCY Zara~\cite{lerner:etal:07} is a publicly available dataset of pedestrians walking in a crowded street near a store, made of CCTV video streams and 489 trajectories (sequences of coordinates), already identified from those images. We used 391 examples (80\%) for training and saved 98 (20\%) for testing. To run both our approaches and the baseline, we first adapted it to the goal recognition task by extracting a map from the video and determining the five main goals reached by those individual agents (store, left street, top right street, right street, bottom right street) from their last seen positions. Figure \ref{fig:ucyimage}(a) displays the five goals (in our experiments, we considered the centroid of each area to be the goal position) and figure \ref{fig:ucyimage}(b) shows the environment extracted from the video, along with an example of a path from a real person. It is clear that captured behaviors do not follow optimal navigation patterns. Moreover, the obstacles may be incorrect, which would challenge the robustness of every approach.

\begin{figure*}[htb]
    \centering
    \includegraphics[width=0.7\linewidth]{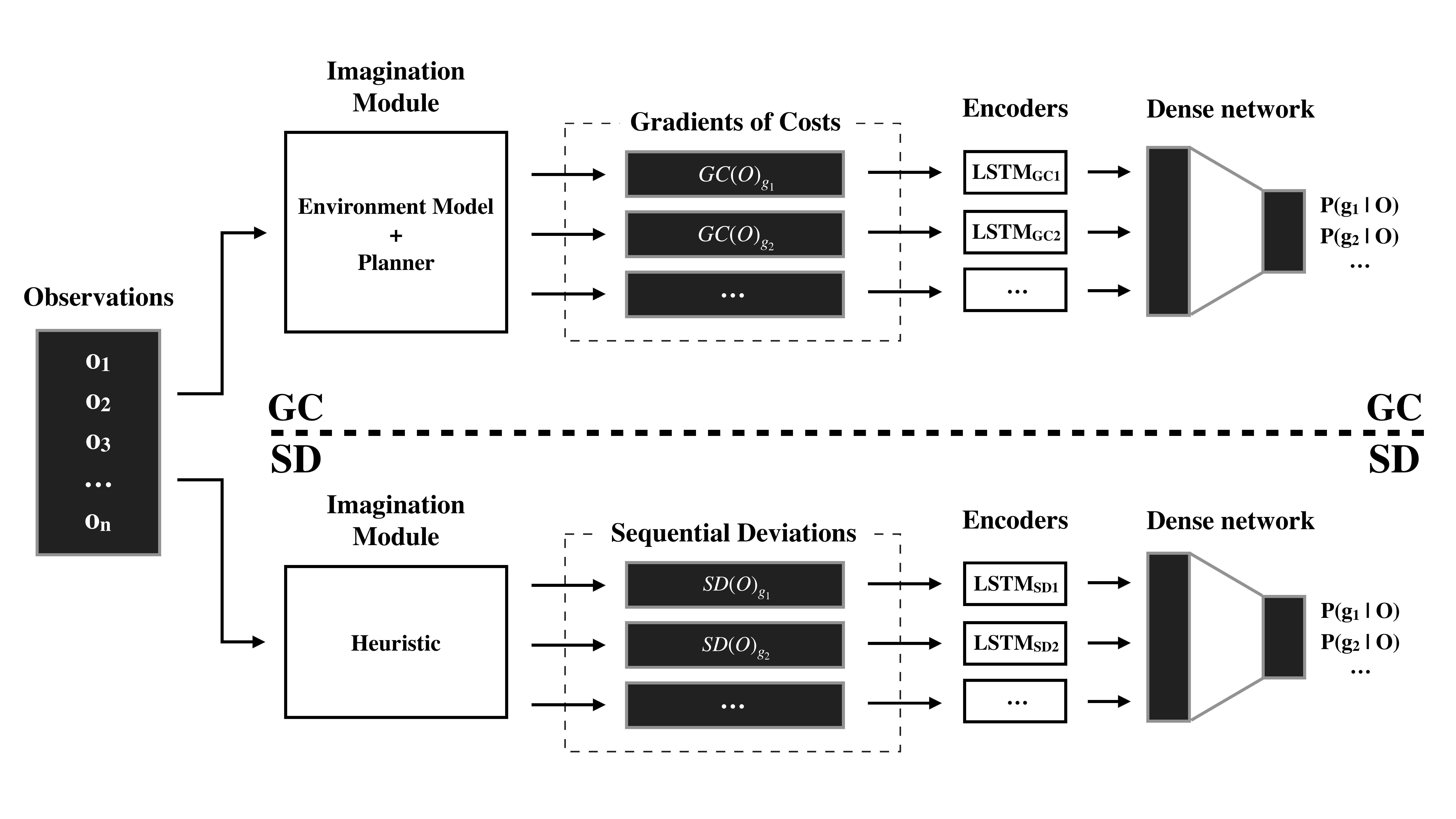}
    \caption{GC (top-half) and SD (bottom-half) learning architectures for the UCY dataset.}
    \label{fig:architecture}
\end{figure*}

While optimal plans were computed using the $A^*$ algorithm for GC and MS~\cite{masters:sardina:19a}, costs were estimated with $h=L_2$ for SD. We used the same learning architecture and hyperparameters for both methods: as for the structure, depicted in figure \ref{fig:architecture}, the encoders are LSTM networks with 64 units each and the dense layer outputs one unit per goal. All the weights are initialized using a uniform \textit{He} distribution~\cite{he:etal:15} and the output is softmax-activated. Since we solve a classification problem, we use the cross-entropy loss function to optimize our network with the Adam algorithm \cite{kingma:ba:14}, whose learning rate is set to 0.001, $\beta_1=0.9$ and $\beta_2=0.999$.

We also built a simple LSTM, using the same initializer and optimizer, to compare the performances of pure deep learning and deep learning augmented with imagination capabilities.

We provide experimental comparisons using the accuracy metric, which is the number of correct predictions over the total number of predictions. A prediction is said to be correct if its highest score corresponds to the ground truth goal. In case of ties, we randomly draw one of the highest scores.

We evaluate the methods at different observable points in time (25\%, 50\%, 75\%, and 100\%). We implement this by truncating our observed paths to the given percentage (for instance, with a path made of 100 observations points and an observability of 25\%, only the first 25 steps are considered).

\begin{figure}[bht]
    \centering
    \includegraphics[width=0.8\linewidth]{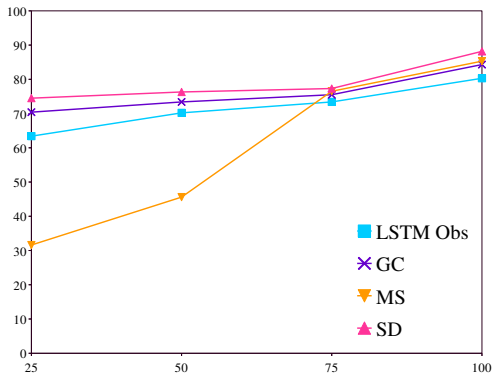}
    \caption{Results on the UCY Zara dataset.}
    \label{fig:resultsUCY}
\end{figure}

Results are shown in figure \ref{fig:resultsUCY} and confirm our hypothesis. We trained a simple LSTM on sequences of coordinates (\textit{LSTM obs} in the graph) to assess how the imagination capability contributes to our deep learning methods. First, the deep learning part of our architecture indeed takes into account apparently erratic behaviors, because we do not assume the level of rationality of the agents. Second, the model extracted from the videos may be incomplete or incorrect, which suggests our approaches are more resilient to erroneous environment knowledge. Finally, the imagination capability helps to improve the performance of GC and SD, compared to a simple deep learning pipeline.

\subsection{Arbitrarily Complex Navigation}

The problem we solve herein is the one of an agent navigating in a grid-world, a benchmark currently used in the state-of-the-art literature \cite{masters:sardina:19a}. It consists in 30 StarCraft maps from the MovingAI Lab website\footnote{https://movingai.com/benchmarks/sc1/index.html} \cite{sturtevant:etal:2012} adapted for goal recognition purposes. The objective is here to infer the destination of an agent by observing a trajectory of their contiguous visited positions. There are four possible actions: move up, down, right, or left. We generated five random goals per map and downscaled them to different sizes to evaluate how the methods would perform on problems of increasing complexity. Though synthetic, we introduced suboptimality in the agent's behavior by generating its path with a modified version of A* that may drop an optimal step and pick a non-optimal one with some chance, using what we define as an $\epsilon$-over-estimating heuristic:
\begin{definition}
An \textit{$\epsilon$-over-estimating heuristic} is a function that returns an admissible quantity $h'$ with a chance of $1-\epsilon$, and $h' + \delta$ otherwise, where $\epsilon \in$ [0, 1] and $\delta > 0$.
\label{def:epsheuristic}
\end{definition}



It is crucial to note that our approaches are trained on a set of different map configurations (obstacles, start, goals) and tested on another set of different map configurations, never seen before, to show the generalization capability of our methods.

\begin{figure*}[htb]
    \centering
    \includegraphics[width=0.8\textwidth]{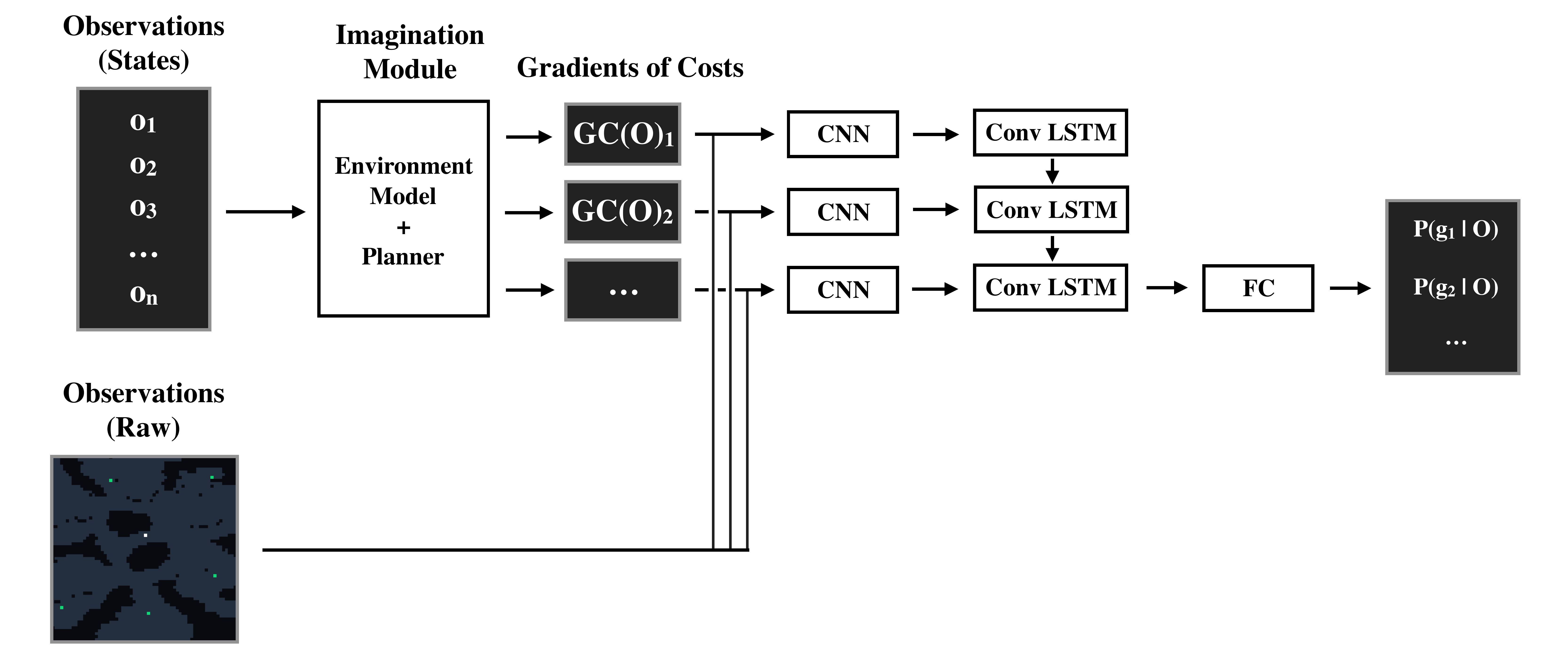}
    \caption{GC network architecture for the navigation domain. The optional max-pooling layers are not displayed.}
    \label{fig:network}
\end{figure*}
We trained the GC-augmented network on full grid observations. Each of the observations takes the form of an 8-channel bitmap bird view of the environment where each channel represents whether the grid cell is an obstacle, a walkable cell, the observed agent's position, or one of its possible goal destinations. Each goal is attributed to a different channel to make them distinctive from one another. We here provided GC features in the form of differential cost maps, with partial derivatives yielded for every position. The resulting matrix was concatenated with the last observation, making it a 9-channel image input.

Since the set of environments used was known and finite, it was possible to compute cost maps offline and to store them before training and testing. To do so, we passed the bitmaps to the breadth-first search (\textit{BFS}) algorithm for every position to generate the remaining cost from them. The resulting cost maps were stored in 30 4-dimension tensors, where the axes represent the coordinates of the start and end positions. The process was repeated for all problem sizes.

The neural network architecture (figure \ref{fig:network}) is composed of:
\begin{enumerate}
    \item 3 convolutional layers (\textit{CNN}) of 16, 32, and 64 3x3 filters respectively, with a stride of one, same padding, and each followed by a ReLU activation;
    \item an optional 2x2 max-pooling layer for 64x64 problems in-between each convolutional layer;
    \item a convolutional LSTM layer (\textit{Conv LSTM}) consisting of 32 3x3 filters for the cell state;
    \item a fully connected layer (\textit{FC}) of 256 units over the flattened output of the LSTM cell;
    \item a final densely connected layer of 5 units followed by a softmax activation for goal inference.
\end{enumerate}

Dropout \cite{srivastava:etal:14} with a drop rate of 0.1 was applied in-between each parametrized layer. The network was trained using the categorical cross-entropy loss for 400 epochs for 16x16 maps and 2000 epochs for 64x64 maps. Each epoch consists of 64 training iterations of mini-batches of size 32. For this benchmark, we generated the examples in parallel to the training process, so that the network may have never seen the same example twice (even in-between epochs). The validation and test sets consist of 160 and 3000 generated examples, respectively.


Finally, the network was optimized using the same initializer and optimizer as in the previous benchmark. Furthermore, the learning rate is gradually reduced by a factor of 0.9 every 10 epochs when a plateau in validation loss is detected, to a minimal value of 1e-5.

As for SD, we trained the same architecture described for the real-world benchmark, for 10 epochs and with 10 000 examples, also generated in parallel to the training process.

\subsubsection{Basic Experiment}

\begin{figure}[bht]
    \centering
    \includegraphics[width=0.8\linewidth]{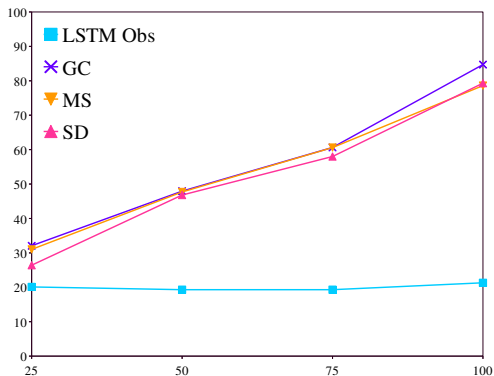}
    \caption{Results on 16x16 grids.}
    \label{fig:results16x16}
\end{figure}

\begin{figure}[bht]
    \centering
    \includegraphics[width=0.8\linewidth]{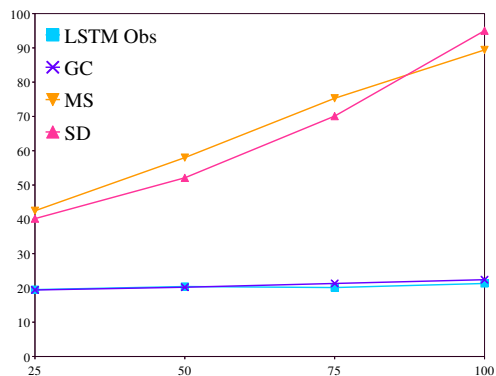}
    \caption{Results on 64x64 grids.}
    \label{fig:results64x64}
\end{figure}

\begin{figure}[htb]
    \centering
    \includegraphics[width=0.8\linewidth]{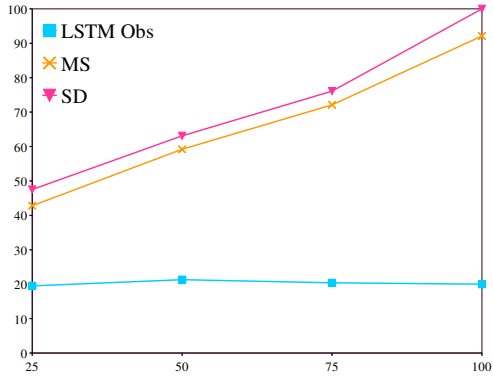}
    \caption{Results on 128x128 grids.}
    \label{fig:results128x128}
\end{figure}

We first tested our methods on different classic problems with accurate models of the environments. Results for grids of size 16x16, 64x64 and 128x128 are shown respectively in figures \ref{fig:results16x16}, \ref{fig:results64x64} and \ref{fig:results128x128}, for $\epsilon=0.2$ and $\delta \in [0,10]$. For small-sized problems, GC outperforms the state-of-the-art algorithm from \citet{masters:sardina:19a} (MS), but SD demonstrates lower accuracy values. Those may be due to complex configurations where paths to different goals are overlapping each other and where approximated metrics such as heuristics cannot fully explain the observed behaviors. However, GC is unable to scale efficiently to larger complexities, as seen in figure~\ref{fig:results64x64}. We can explain this phenomenon by considering the rapidly increasing input size for longer paths. Indeed, a sequence of 64x64 observations is eight times bigger than the same sequence of 16x16 ones. The network architecture was not enough complex to learn with such a rich input, despite the max-pooling layers reducing its dimension. As a result, we could not train GC on 128x128 maps with our available computing resources. On the other hand, SD surpasses other techniques when given larger problems, since it may convey more information about the general temporal moving direction when given longer sequences.

We may explain this outcome by reasoning about the amount of information we provide to each method. Gradients of costs embed every single movement, which is why they produce precise predictions but are expensive to compute. Symbolic algorithms, to the contrary, are limited only to two costs per goal and therefore deprived of heavily cutoff information in the temporal dimension. Finally, sequential deviations seem to withhold sufficient clues about the temporal sequences, without needing to compute precise costs.

\subsubsection{Robustness to Erroneous Models}

We then experimented with erroneous representations of the environments. To implement this notion, we applied definition~\ref{def:epsheuristic} to modify the costs of transitions when computing the required paths for both GC (in the future module) and \citet{masters:sardina:19a} (in formula~\ref{costdiff:ms}), such that we add a random value $\delta' \in [0,10]$ to the real transition cost with a chance of $\epsilon' \in [0,1]$. SD, whose computation does not require planners (and thus no specific domain knowledge), is not affected by this process.

\begin{figure}[bht]
    \centering
    \includegraphics[width=0.8\linewidth]{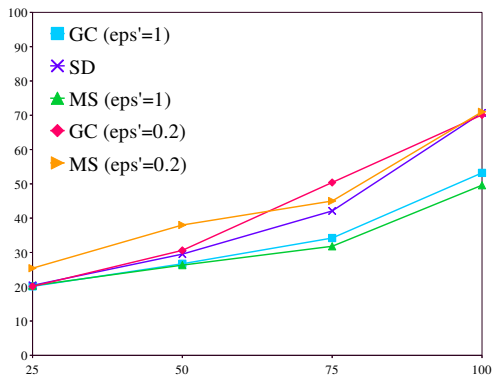}
    \caption{Results for $\epsilon' = 0.2$ and $\epsilon' = 1$ (16x16 grid).}
    \label{fig:robustness}
\end{figure}

Results are shown in figure \ref{fig:robustness} for $\epsilon' = 0.2$ and $\epsilon' = 1$ and illustrate that our methods handle incorrect environments more efficiently. We believe that the learning part of our model is crucial to adapt to such misbeliefs. It is also interesting to note how heuristics are unconditional estimates that do not essentially depend on environment knowledge.

\section{Related Work}

Deep learning has proven more than efficient for unstructured data classification, such as images and raw sensor data. Consequently, numerous architectures were experimented for short-term activity analysis \cite{liu:etal:18,kong:fu:18}. However, they mainly focus on identifying immediate actions without considering a larger temporal scope.

Although deep learning has made tremendous inroads in various activity recognition domains, it is surprisingly underused for agents engaged in long-term planning processes. Only a few research works explored the horizon of recurrent networks for long-term goal recognition. \citet{min:etal:16} made use of LSTM networks to recognize the goal of a player from sequences of interactions in the game of \textsc{Crystal Island}, displaying promising results. \citet{amado:etal:18} assembled a working pipeline with existing tools (a dense auto-encoder network from \citet{asai:etal:18} with an LSTM), casting the problem of goal recognition as a regression task in a latent space for small games like 8-puzzle or tower of Hanoi.

An attractive alternative for goal or plan recognition that we exploit ourselves lies in the combination of the learning paradigm with a symbolic one to automatically approximate some domain knowledge from observations that complement expert resources. \citet{bisson:etal:15} created a deep architecture mimicking HTN plan libraries to tune probabilistic inference models for plan recognition automatically. \citet{granada:etal:17} built a convolutional network to identify primitive actions from videos and combined it with the Symbolic Behavior Recognition (SBR) algorithm based on an HTN plan library to detect the goals achieved in a kitchen environment. \citet{pereira:etal:19} introduced a manner to construct a nominal model of the environment (states and transition rules) by the use of a deep network and then perform goal recognition using a cost-based inference algorithm. The difference between these works and ours is that our models learn directly from knowledge whose format is non-specific to any environment, making them transferable to multiple ones.


The future projection capability is seeing a growing research interest from the deep learning community. Imagination-augmented agents~\cite{racaniere:etal:17} that inspired our work go further by using model-based deep reinforcement learning ideas to imagine future projected trajectories to guide the exploration of a model-free deep policy learner in Sokoban, PacMan and other related games. \citet{dosovitskiy:koltun:17} transform the standard reinforcement learning setting into a self-supervised one by attempting to predict action effects on measurements (such as altitude, health). \citet{ha:schmidhuber:18} use variational auto-encoders to simulate world models from games and an evolutionary algorithm to learn from these simulations. \citet{ke:etal:19} effectively learn to predict some long-term future using improved LSTM architectures and show how it helps in various planning tasks, either deep-learned by imitation or by reinforcement. While these approaches performed on multiple problems involving long-term reasoning, long-term goal recognition is not one of them. Another aspect is that they all chose to learn future projection, while we rely on symbolic models and planners. They enabled us to achieve impressive results on challenging problems using a simpler architecture and fewer data.


\section{Conclusion}

We presented two innovative solutions to goal recognition by combining imagination-augmented deep learning architectures with costs-based features derived from symbolic knowledge. The ability to project the observed agent into the future, which is inherent to long-term goal recognition, helps generalize to multiple configurations. The first metric, \textit{gradients of costs}, encodes more temporal information that just a difference of costs, but is expensive. The second one, \textit{sequential deviations}, helps reduce the computational cost by approximating the previous one with heuristic functions so that no planner is required anymore.

Our solution outperforms the state-of-the-art sheer symbolic methods, both in synthetic and real environments. We demonstrated that our approaches could more efficiently predict the goal of the agent when our assumptions about their behavior are wrong (that is, when they are suboptimal and when the environment model is erroneous), hence proving its robustness.

\section{Acknowledgements}

We would like to thank \textit{Compute Canada} for the computing resources they provided and the \textit{NVIDIA Corporation} for the Quadro P6000 graphic card they donated.

\bibliographystyle{aaai.bst}  

\bibliography{bibliography}  

\begin{thebibliography}{}

\bibitem[\protect\citeauthoryear{Amado \bgroup et al\mbox.\egroup
  }{2018}]{amado:etal:18}
Amado, L.; Aires, J.~P.; Pereira, R.~F.; Magnaguagno, M.~C.; Granada, R.; and
  Meneguzzi, F.
\newblock 2018.
\newblock L{STM}-based goal recognition in latent space.
\newblock {\em CoRR} abs/1808.05249.

\bibitem[\protect\citeauthoryear{Asai and Fukunaga}{2018}]{asai:etal:18}
Asai, M., and Fukunaga, A.
\newblock 2018.
\newblock Classical planning in deep latent space: Bridging the
  subsymbolic-symbolic boundary.
\newblock {\em AAAI}.

\bibitem[\protect\citeauthoryear{Bisson, Larochelle, and
  Kabanza}{2015}]{bisson:etal:15}
Bisson, F.; Larochelle, H.; and Kabanza, F.
\newblock 2015.
\newblock Using a recursive neural network to learn an agent's decision model
  for plan recognition.
\newblock In {\em {IJCAI} 2015},  918--924.

\bibitem[\protect\citeauthoryear{Bylander}{1994}]{bylander:94}
Bylander, T.
\newblock 1994.
\newblock The computational complexity of propositional strips planning.
\newblock {\em Artificial Intelligence} 69(1):165 -- 204.

\bibitem[\protect\citeauthoryear{Dosovitskiy and
  Koltun}{2017}]{dosovitskiy:koltun:17}
Dosovitskiy, A., and Koltun, V.
\newblock 2017.
\newblock Learning to act by predicting the future.
\newblock In {\em {ICLR} 2017}.

\bibitem[\protect\citeauthoryear{E.{-}Mart{\'{\i}}n, R.{-}Moreno, and
  Smith}{2015}]{e-martin:etal:15}
E.{-}Mart{\'{\i}}n, Y.; R.{-}Moreno, M.~D.; and Smith, D.~E.
\newblock 2015.
\newblock A fast goal recognition technique based on interaction estimates.
\newblock In {\em {IJCAI} 2015},  761--768.
\newblock Buenos Aires, Argentina: {AAAI} Press.

\bibitem[\protect\citeauthoryear{Granada \bgroup et al\mbox.\egroup
  }{2017}]{granada:etal:17}
Granada, R.; Pereira, R.; Monteiro, J.; Barros, R.; Ruiz, D.; and Meneguzzi, F.
\newblock 2017.
\newblock Hybrid activity and plan recognition for video streams.
\newblock In {\em {AAAI} 2017}.

\bibitem[\protect\citeauthoryear{Ha and Schmidhuber}{2018}]{ha:schmidhuber:18}
Ha, D., and Schmidhuber, J.
\newblock 2018.
\newblock Recurrent world models facilitate policy evolution.
\newblock In {\em NeurIPS 2018},  2450--2462.

\bibitem[\protect\citeauthoryear{He \bgroup et al\mbox.\egroup
  }{2015}]{he:etal:15}
He, K.; Zhang, X.; Ren, S.; and Sun, J.
\newblock 2015.
\newblock Delving deep into rectifiers: Surpassing human-level performance on
  imagenet classification.
\newblock {\em CoRR} abs/1502.01852.

\bibitem[\protect\citeauthoryear{Ke \bgroup et al\mbox.\egroup
  }{2019}]{ke:etal:19}
Ke, N.~R.; Singh, A.; Touati, A.; Goyal, A.; Bengio, Y.; Parikh, D.; and Batra,
  D.
\newblock 2019.
\newblock Modeling the long term future in model-based reinforcement learning.
\newblock In {\em ICLR 2019}.

\bibitem[\protect\citeauthoryear{Kingma and Ba}{2014}]{kingma:ba:14}
Kingma, D.~P., and Ba, J.
\newblock 2014.
\newblock Adam: A method for stochastic optimization.
\newblock {\em CoRR} abs/1412.6980.

\bibitem[\protect\citeauthoryear{Kong and Fu}{2018}]{kong:fu:18}
Kong, Y., and Fu, Y.
\newblock 2018.
\newblock Human action recognition and prediction: {A} survey.
\newblock {\em CoRR} abs/1806.11230.

\bibitem[\protect\citeauthoryear{Lerner, Chrysanthou, and
  Lischinski}{2007}]{lerner:etal:07}
Lerner, A.; Chrysanthou, Y.; and Lischinski, D.
\newblock 2007.
\newblock Crowds by example.
\newblock {\em Comput. Graph. Forum} 26:655--664.

\bibitem[\protect\citeauthoryear{Liu \bgroup et al\mbox.\egroup
  }{2018}]{liu:etal:18}
Liu, K.; Liu, W.; Gan, C.; Tan, M.; and Ma, H.
\newblock 2018.
\newblock T-c3d: Temporal convolutional 3d network for real-time action
  recognition.
\newblock {\em AAAI}.

\bibitem[\protect\citeauthoryear{Masters and
  Sardi{\~{n}}a}{2019a}]{masters:sardina:19a}
Masters, P., and Sardi{\~{n}}a, S.
\newblock 2019a.
\newblock Cost-based goal recognition in navigational domains.
\newblock {\em JAIR} 64:197--242.

\bibitem[\protect\citeauthoryear{Masters and
  Sardina}{2019b}]{masters:sardina:19b}
Masters, P., and Sardina, S.
\newblock 2019b.
\newblock Goal recognition for rational and irrational agents.
\newblock In {\em AAMAS 2019},  440--448.

\bibitem[\protect\citeauthoryear{Min \bgroup et al\mbox.\egroup
  }{2016}]{min:etal:16}
Min, W.; Mott, B.; Rowe, J.; Liu, B.; and Lester, J.
\newblock 2016.
\newblock Player goal recognition in open-world digital games with long
  short-term memory networks.
\newblock In {\em {IJCAI} 2016},  2590--2596.
\newblock AAAI Press.

\bibitem[\protect\citeauthoryear{Pereira \bgroup et al\mbox.\egroup
  }{2019}]{pereira:etal:19}
Pereira, R.~F.; Vered, M.; Meneguzzi, F.; and Ram{\'{\i}}rez, M.
\newblock 2019.
\newblock Online probabilistic goal recognition over nominal models.
\newblock In {\em Proceedings of the Twenty-Eighth International Joint
  Conference on Artificial Intelligence, {IJCAI} 2019, Macao, China, August
  10-16, 2019},  5547--5553.

\bibitem[\protect\citeauthoryear{Pereira, Oren, and
  Meneguzzi}{2017}]{pereira:etal:17}
Pereira, R.~F.; Oren, N.; and Meneguzzi, F.
\newblock 2017.
\newblock Landmark-based heuristics for goal recognition.
\newblock In {\em Proceedings of the Thirty-First {AAAI} Conference on
  Artificial Intelligence},  3622--3628.
\newblock San Francisco, California, {USA}: {AAAI} Press.

\bibitem[\protect\citeauthoryear{Racani{\`{e}}re \bgroup et al\mbox.\egroup
  }{2017}]{racaniere:etal:17}
Racani{\`{e}}re, S.; Weber, T.; Reichert, D.~P.; Buesing, L.; Guez, A.;
  Rezende, D.~J.; Badia, A.~P.; Vinyals, O.; Heess, N.; Li, Y.; Pascanu, R.;
  Battaglia, P.~W.; Hassabis, D.; Silver, D.; and Wierstra, D.
\newblock 2017.
\newblock Imagination-augmented agents for deep reinforcement learning.
\newblock In {\em NIPS 2017},  5690--5701.

\bibitem[\protect\citeauthoryear{Ram{\'{\i}}rez and
  Geffner}{2009}]{ramirez:geffner:09}
Ram{\'{\i}}rez, M., and Geffner, H.
\newblock 2009.
\newblock Plan recognition as planning.
\newblock In {\em Proceedings of the 21st International Joint Conference on
  Artificial Intelligence},  1778--1783.
\newblock Pasadena, California, USA: {AAAI} Press.

\bibitem[\protect\citeauthoryear{Ram\'{\i}rez and
  Geffner}{2010}]{ramirez:geffner:10}
Ram\'{\i}rez, M., and Geffner, H.
\newblock 2010.
\newblock Probabilistic plan recognition using off-the-shelf classical
  planners.
\newblock In {\em AAAI 2010},  1121--1126.
\newblock Atlanta, Georgia: AAAI Press.

\bibitem[\protect\citeauthoryear{Sohrabi, Riabov, and
  Udrea}{2016}]{sohrabi:etal:16}
Sohrabi, S.; Riabov, A.~V.; and Udrea, O.
\newblock 2016.
\newblock Plan recognition as planning revisited.
\newblock In {\em Proceedings of the Twenty-Fifth International Joint
  Conference on Artificial Intelligence}, IJCAI'16,  3258--3264.
\newblock AAAI Press.

\bibitem[\protect\citeauthoryear{Srivastava \bgroup et al\mbox.\egroup
  }{2014}]{srivastava:etal:14}
Srivastava, N.; Hinton, G.~E.; Krizhevsky, A.; Sutskever, I.; and
  Salakhutdinov, R.
\newblock 2014.
\newblock Dropout: a simple way to prevent neural networks from overfitting.
\newblock {\em Journal of Machine Learning Research} 15(1):1929--1958.

\bibitem[\protect\citeauthoryear{Sturtevant}{2012}]{sturtevant:etal:2012}
Sturtevant, N.
\newblock 2012.
\newblock Benchmarks for grid-based pathfinding.
\newblock {\em Transactions on Computational Intelligence and AI in Games}
  4(2):144 -- 148.

\bibitem[\protect\citeauthoryear{Vered and Kaminka}{2017}]{vered:kaminka:17}
Vered, M., and Kaminka, G.~A.
\newblock 2017.
\newblock Heuristic online goal recognition in continuous domains.
\newblock In {\em Proceedings of the Twenty-Sixth International Joint
  Conference on Artificial Intelligence, {IJCAI} 2017},  4447--4454.
\newblock Melbourne, Australia: ijcai.org.

\bibitem[\protect\citeauthoryear{Vered, Kaminka, and
  Biham}{2016}]{vered:etal:16}
Vered, M.; Kaminka, G.; and Biham, S.
\newblock 2016.
\newblock Online goal recognition through mirroring: humans and agents.
\newblock In {\em Fourth Annual Conference on Advances in Cognitive Systems},
  Advances in Cognitive Systems.
\newblock Evanston, Illinois, (USA): Cognitive Systems Foundation.

\end{thebibliography}

\end{document}